\documentclass[letterpaper]{article} 
\usepackage{aaai25}  
\usepackage{times}  
\usepackage{helvet}  
\usepackage{courier}  
\usepackage[hyphens]{url}  
\usepackage{graphicx} 
\usepackage{natbib}  
\usepackage{caption} 
\usepackage{algorithm}
\usepackage{algorithmic}
\usepackage{subcaption}
\usepackage{amsmath,amssymb, bm}
\usepackage{multirow}
\usepackage{newfloat}
\usepackage{listings}
\DeclareCaptionStyle{ruled}{labelfont=normalfont,labelsep=colon,strut=off} 
\floatstyle{ruled}
\newfloat{listing}{tb}{lst}{}
\floatname{listing}{Listing}
\title{Self-attention-based Diffusion Model for Time-series Imputation \\ in Partial Blackout Scenarios}
\author {
    Mohammad Rafid Ul Islam, 
    Prasad Tadepalli, 
    Alan Fern
}
\affiliations {
    Oregon State University\\
    islammoh@oregonstate.edu, prasad.tadepalli@oregonstate.edu, alan.fern@oregonstate.edu
}
\usepackage{bibentry}
\begin{document}

\maketitle

\begin{abstract}
Missing values in multivariate time series data can harm machine learning performance and introduce bias. These gaps arise from sensor malfunctions, blackouts, and human error and are typically addressed by data imputation. Previous work has tackled the imputation of missing data in random, complete blackouts and forecasting scenarios. The current paper addresses a more general missing pattern, which we call ``partial blackout,'' where a subset of features is missing for consecutive time steps. We introduce a two-stage imputation process using self-attention and diffusion processes to model feature and temporal correlations. Notably, our model effectively handles missing data during training, enhancing adaptability and ensuring reliable imputation and performance, even with incomplete datasets. 
Our experiments on benchmark and two real-world time series datasets demonstrate that our model outperforms the state-of-the-art in partial blackout scenarios and shows better scalability. 
\end{abstract}

\section{Introduction}
\label{sec:intro}
Multivariate time-series data is common in many fields like finance, meteorology, agriculture, transportation, and healthcare. The data often has  missing values 
due to human error, equipment failure, or bad data entry \citep{Silva2012-cb, geo-sensory-miss}.
Most machine learning algorithms 
perform poorly when data are incomplete. 
Poor imputations can degrade task quality and introduce bias, compromising integrity of the result \citep{shadbahr2022classification, Zhang2021-pw}.


There have been numerous 
studies of time series data imputation in both statistical learning and deep learning 
communities. The statistical learning techniques include mean/median imputation, linear interpolation, K-nearest neighbor imputation, 
and more advanced iterative regression-based models, such as MICE \cite{mice}. 
More recent deep neural network-based models include 
autoregressive models, such as, GRU-D \cite{gru-d}, BRITS \citep{brits}; and self-attention-based models, such as, SAITS \citep{saits}. 
Generative models have also been explored, including GAN-based approaches \cite{e2gan, grui-gan, naomi, miao2021generative} and VAE-based methods \cite{gp-vae}, which bring additional training-stability challenges. 

In recent years, score-based diffusion models have 
gained prominence in domains such as
image generation \citep{ho2020denoising, nichol2021improved, ode-score, 3daware} and audio synthesis \citep{diffwave, wavegrad}, and 
have been applied to imputation and forecasting in time series data.  
Two such methods, CSDI \cite{csdi} and SSSD \cite{sssd}, in particular,  provide good quality imputations by conditioning on observed values. Another diffusion model, Time-Grad \cite{time-grad-25}, excels in forecasting but can't leverage future data for imputation. Bayesian inference models like BGCF \cite{Cui2019}, \cite{Vidotto2018}, and \cite{Vidotto2019} address mixed and longitudinal data. Other methods include graph neural networks like GRIN \citep{grin}, which require high sensor homogeneity, and latent ODE networks with RNNs \citep{latent-ode}, which improve modeling of irregularly sampled time series but demand more computational resources. Another approach \citep{long-term-Park2022} employs hyper-parameter optimization for an MLP architecture for long-term imputation, but faces challenges in accurately estimating end-of-gap values.


A recurring theme in many of these works is the attention to limited types 
of missingness. They mostly focus on 
randomly missing data, missing a feature for a few consecutive time steps (interpolation), and complete blackouts where all features are missing for some time. 
In our study, we explore a more general and common type of missingness we call ``partial blackout,'' 
which covers situations where a variable number of features become unavailable for some time. 

\begin{figure}[h!]
    \centering
   
        \includegraphics[width=0.35\textwidth]{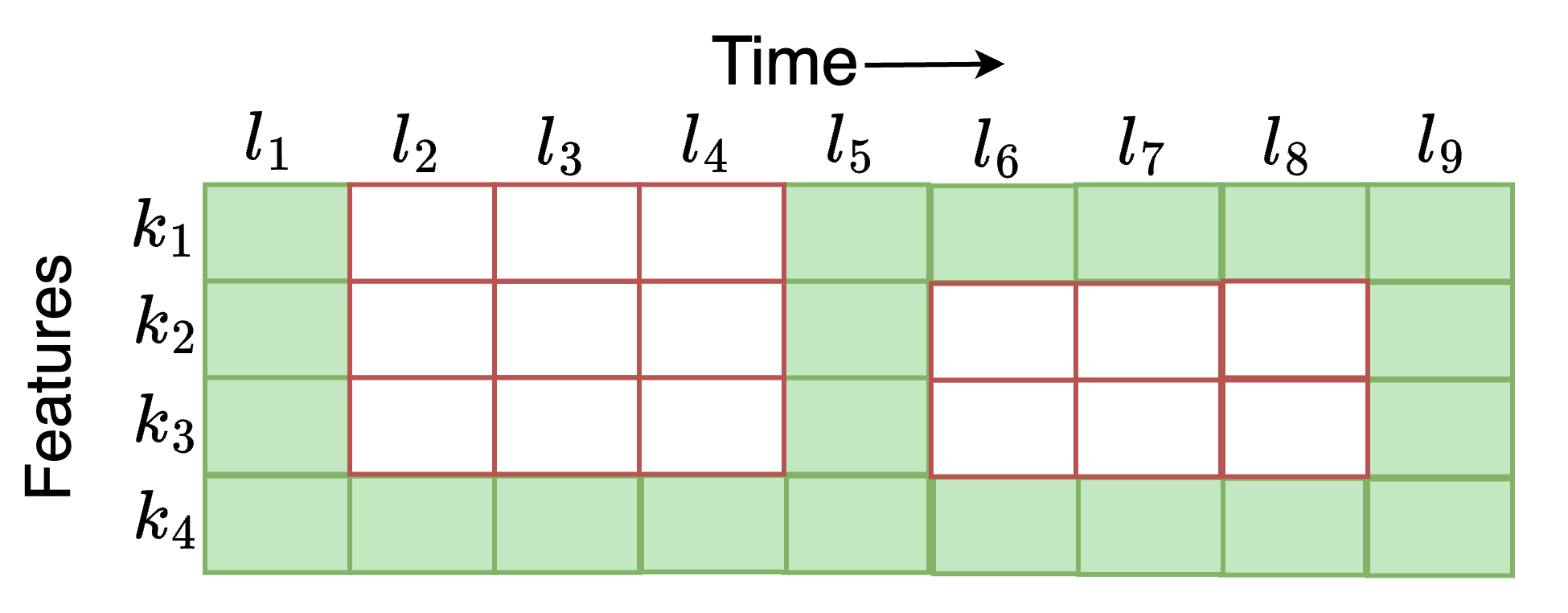}

    \caption{Partial blackouts.  Multiple features are missing in consecutive time steps as shown by the empty cells.}
    \label{fig:pbm}
\end{figure}

Importantly, this concept 
allows multiple distinct blocks of missingness, each characterized by a set of 
absent features for a number of steps 
as shown in Figure ~\ref{fig:pbm}. The random missing scenario 
can be expressed as a partial blackout case where multiple single feature blocks are missing for one time step. Similarly, the interpolation 
case can be expressed as a partial blackout with one feature missing for a period of time. 
Scenarios such as complete blackout   
and forecasting
 are just extreme cases of partial blackouts. 
 Thus the category of partial blackouts allows us to address a broad spectrum of missingness patterns in real-world datasets.

In the current work, 
we tackle the challenge of 
partial blackout
 problem in time series imputation. 
We introduce the \emph{Self-Attention Diffusion Model for Time Series Imputation (SADI)} as a novel generative model for diffusion-based imputation.
The core components of our model are the explicit modeling of 
feature dependencies across time in the form of a \emph{Feature Dependency Encoder} and the modeling of temporal correlations across features in the form of \emph{Gated Temporal Attention} blocks. We also deploy a two-stage imputation process 
where the imputed data from the initial stage is further refined and impacts the imputation of other data in the second stage. 
The model combines the results of the two stages in a weighted combination where the weights are learned from the data. Our model is designed to operate without the assumption of complete data during the training process. Importantly, our training approach is able to learn from data with missing values by utilizing a masking strategy that allows it to track the data that is originally missing. In summary, our contributions are threefold. 

\begin{enumerate}
    \item \textbf{Exploring partial blackout scenarios}: We introduce a new archetype of missing patterns called partial blackout, a more generalized concept encompassing the limited cases of missingness investigated in previous studies. 
    \item \textbf{Explicit modeling of feature and temporal correlations and two-stage imputation}: We introduce a novel diffusion-based imputation model, SADI, which captures feature and time dependencies with explicit components within the denoising function. The model includes 
    a two-stage imputation process, where the second stage refines the first stage imputations, improving the overall quality.

    \item \textbf{Empirical study}: We evaluate our model in several real-world time series domains 
    with randomly missing training data 
    and show that it outperforms and scales better than the 
    other state-of-the-art models under partial blackout scenarios. It 
    improves further by fine tuning with partial blackout training mixed with randomly missing data.   

\end{enumerate}



\section{Diffusion Models for Imputation}
\label{sec:background}
Diffusion-based probabilistic models have two processes: forward and reverse. The forward process adds noise to the original data incrementally until it becomes pure noise. The reverse process then removes the noise step-by-step, starting from pure noise and reconstructing the original data distribution over the same number of iterations \cite{sohl2015deep, ho2020denoising}.

A diffusion model involves approximating a data distribution $q(X_0)$ by learning a model distribution $p_{\theta}(X_0)$. Let $X_t$ for $t = \{1,\ldots, T\}$ be the latent variables representing the noisy data at diffusion step $t$. These belong to the same sample space as the original data, $X_0$. The forward diffusion process, which is a Markov chain is defined by

\[q(X_{1:T}|X_0) = \prod_{t=1}^T q(X_t|X_{t-1}) \]

\begin{equation}
    \label{eq:forward}
    q(X_t|X_{t-1})=\mathcal{N}(\sqrt{1-\beta_t}X_{t-1},\beta_t\mathbf{I})
\end{equation}
Here, $\beta_t$ represents the variance of the noise applied at each diffusion step $t$ of the forward process. Furthermore, $X_t$ has a closed form $X_t=\sqrt{\bar{\alpha}_t}X_0+\sqrt{(1-\bar{\alpha}_t)}\epsilon$, where $\alpha_t=1-\beta_t$, $\bar{\alpha}_t = \prod_{i=1}^t\alpha_i$, and $\epsilon \sim \mathcal{N}(0,\mathbf{I})$.  The reverse process is also a Markov chain that denoises $X_t$ to get $X_{t-1}$ using a denoising function $\epsilon_{\theta}$. After $T$ such iterations, we regenerate $X_0$. The reverse process is defined by:
\[p_{\theta}(X_{0:T}) = p(X_T)\prod_{t=1}^Tp_{\theta}(X_{t-1}|X_t)\mbox{, where } X_T \sim \mathcal{N}(0, \mathbf{I})\]
\begin{equation}
    \label{eq:reverse}
    p_{\theta}(X_{t-1}|X_t)=\mathcal{N}(\mu_{\theta}(X_t,t),\sigma_{\theta}(X_t,t)\mathbf{I})
\end{equation}
Here, $p_{\theta}(X_{t-1}|X_t)$ is a learnable function. Following  \citep{ho2020denoising}, we have:
\begin{equation}
    \label{eq:mean}
    \mu_{\theta}(X_t,t) = \frac{1}{\sqrt{\alpha_t}}\left(X_t - \frac{\beta_t}{\sqrt{1-\bar{\alpha}_t}}\bm{\epsilon_{\theta}}(X_t, t)\right)
\end{equation}
where  $\sigma_{\theta}(X_t,t)$ is kept constant for each diffusion step $t$. In Eq. \ref{eq:mean}, the $\bm{\epsilon_{\theta}}$ denoising function is trainable. Using the parameterization of $\mu_{\theta}(X_t,t)$ in Eq. \ref{eq:mean},  
the reverse process can be trained by optimizing the following objective \cite{ho2020denoising}:
\begin{equation}
    \label{eq:loss}
    L = \min_{\theta}\mathbb{E}_{X\sim q(X_0), \epsilon\sim \mathcal{N}(0,\mathbf{I}),t}||\epsilon-\bm{\epsilon_{\theta}}(X_t,t) ||_2^2
\end{equation}
The denoising function $\bm{\epsilon_{\theta}}$ takes the noisy data at step t, $X_t$, and the diffusion step $t$ as inputs and estimates the added noise $\epsilon$ introduced to the noisy input $X_{t-1}$ to get $X_t$ in the forward diffusion step. Once the training is completed, we can sample $X_0$ by following Eqs.~ \ref{eq:reverse} and \ref{eq:mean}.

Diffusion modeling is applied 
to the 
imputation problem by making the reverse process conditioned on the observed values for the posterior estimation. Given a sample $X_0$, we condition on the observed values, $X_0^{co}$ and generate the imputation targets, $X_0^{ta}$. We extend Eqs. \ref{eq:reverse} and \ref{eq:mean} to account for the conditional input $X_0^{co}$. We now have,
\begin{equation}
    \label{eq:reverse_cond}
    \begin{split}
    p_{\theta}(X^{ta}_{t-1}|X^{ta}_t, X_0^{co}) &=\mathcal{N}(\mu_{\theta}(X^{ta}_t, X_0^{co}, t),\\&\sigma_{\theta}(X^{ta}_t, X_0^{co}, t)\mathbf{I})
    \end{split}
\end{equation}
\begin{equation}
    \label{eq:reverse_mean}
    \begin{split}
        \mu_{\theta}(X_t^{ta}, X_0^{co},t) =& \frac{1}{\sqrt{\alpha_t}}(X^{ta}_t - \frac{\beta_t}{\sqrt{1-\bar{\alpha}_t}}\\&\bm{\epsilon_{\theta}}(X^{ta}_t, X^{co}_0, t))
    \end{split}
\end{equation}
where, $X_t^{ta}$ is the noisy data at step $t$, and $\sigma_{\theta}(X^{ta}_t,  X_0^{co}, t)$ remains the same as in the unconditional case. Importantly, we aim for a model where the sets of observed and unobserved variables can change from instance to instance during training and testing.

\section{Self-Attention Diffusion Model for Time-series Imputation}
\label{sec:methodology}
We consider multivariate time-series data, represented as 
$X_0 = \{x_{1:L,1:K}\} \in \mathbf{R}^{L\times K}$,
where $L$ is the time-series length, $K$ is the number of features, 
and the subscript $0$ indicates the diffusion step.
The data follows the joint distribution $\mathbf{P}(X_0)$, where a part, $X_0^{co}$, is observed, and the rest, $X_0^{ta}$, missing. The imputation problem is to generate the missing data according to $\mathbf{P}(X_0^{ta}|X_0^{co})$. To indicate which parts of the input are observed, we use a binary mask $M_0^{co} = \{m_{1:L,1:K}\} \in \{0, 1\}^{L\times K}$, where $1$ indicates that the data is observed. By convention any unobserved/missing values in $X_0$ are set to zero. 
The training data consists of samples from $\mathbf{P}$ with some values possibly missing. 

\begin{figure}[t!]
    \centering 
    \begin{subfigure}[b]{0.23\textwidth}
        \centering
        \includegraphics[width=\textwidth]{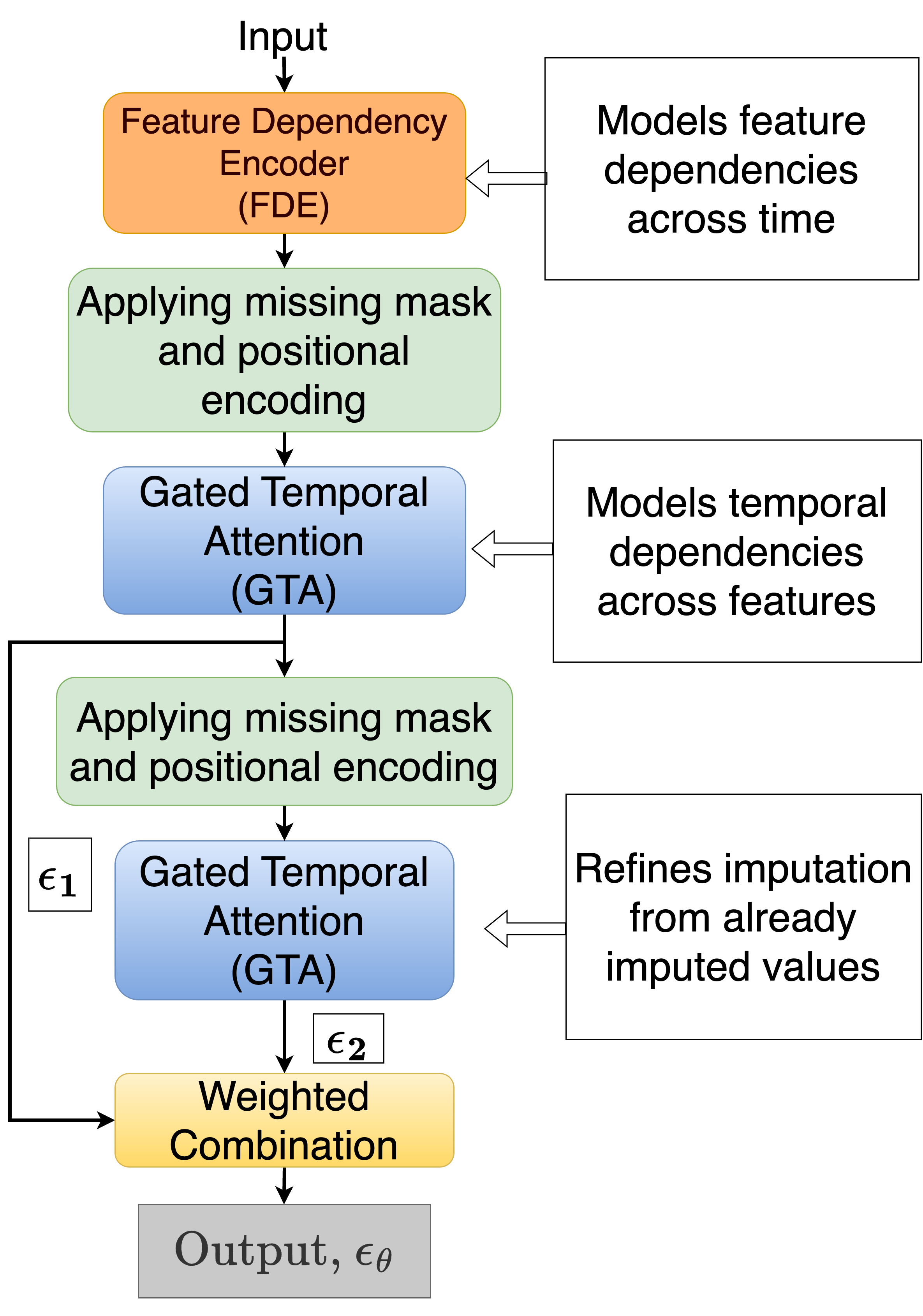}
        \caption{Overview of the architecture of our denoising function, $\epsilon_{\theta}$}
        \label{fig:sadi}
    \end{subfigure}
    \hfill
    \begin{subfigure}[b]{0.23\textwidth}
        \centering
        \includegraphics[width=\textwidth]{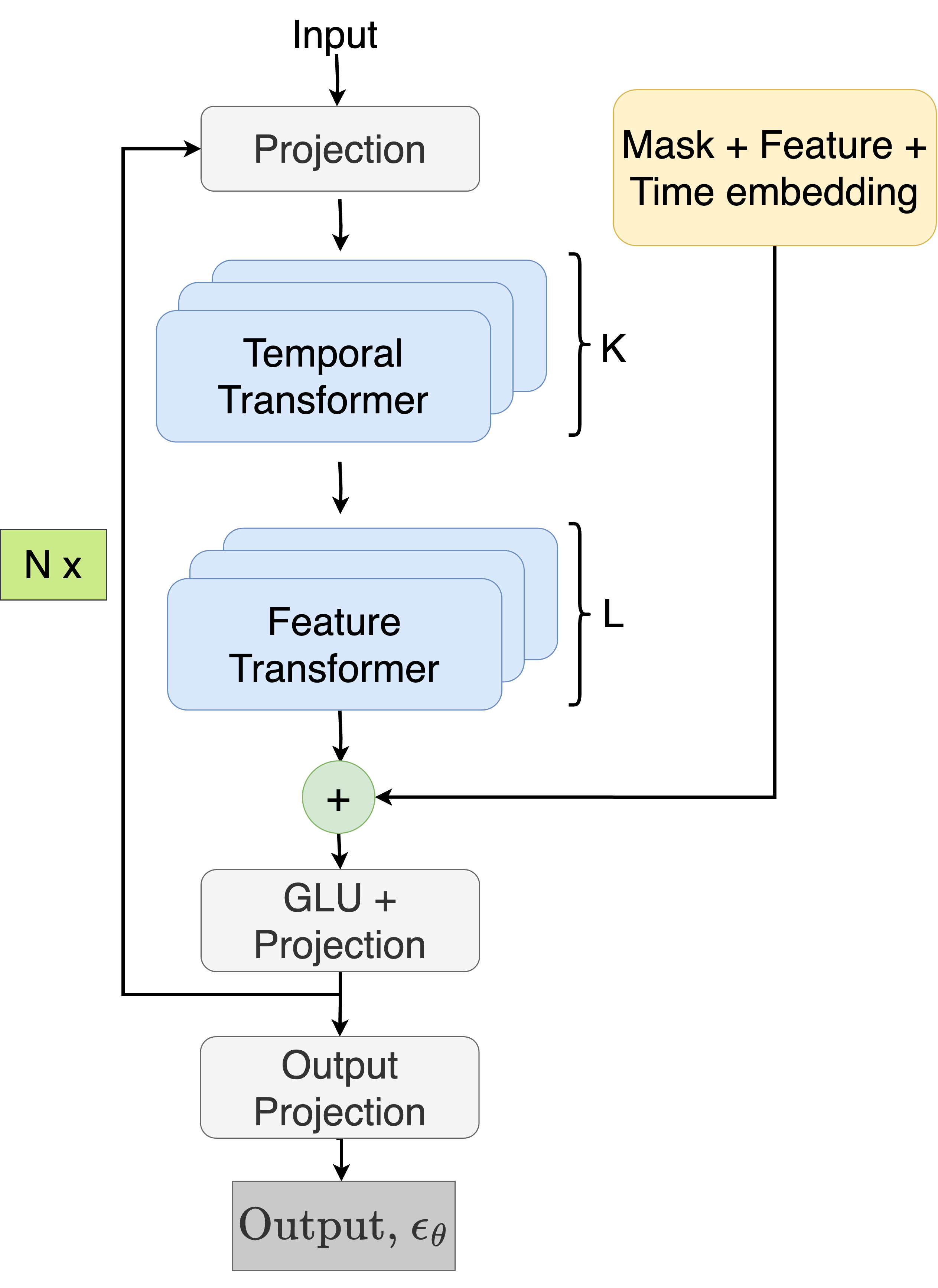}
        \caption{Overview of the architecture of CSDI's $\epsilon_{\theta}$ function}
        \label{fig:csdi}
    \end{subfigure}
    \caption{Architectures of SADI and CSDI models.  CSDI uses consecutive transformers to capture feature and temporal dependencies, partitioning features into $K$ 
    instances and time steps into $L$ 
    instances. In contrast, SADI models feature and time dependencies jointly.}
    \label{fig:overview}
\end{figure}

Similarly to the CSDI model \cite{csdi}, our model SADI leverages the conditional diffusion process framework from \cite{ho2020denoising} for time series imputation, modeling the denoising function $\bm{\epsilon_{\theta}}(X^{ta}_t, X^{co}_0, t)$ as in Eq.\ref{eq:reverse_mean}. While CSDI (Figure\ref{fig:csdi}) uses separate transformers for feature and time correlations, isolating each time step and segmenting features, SADI (Figure~\ref{fig:sadi}) employs a \emph{Feature Dependency Encoder (FDE)} and a \emph{Gated Temporal Attention (GTA)} block, inspired by the SAITS model \cite{saits}, to jointly capture feature and time correlations.

The FDE uses a 1-D convolution and self-attention to model time-aware feature correlations, while the GTA focuses on temporal dependencies. Our two-stage imputation process, inspired by the DMSA blocks in SAITS, combines intermediary imputations using attention maps and missingness information. Unlike CSDI’s back-to-back transformers applied on the multiple channels separating the time and feature dimensions, SADI uses a single-channel architecture with explicit self-attention components. Notably, SADI’s FDE is time-aware in modeling feature dependencies, whereas CSDI treats each time step as i.i.d.

The diffusion denoising function predicts the noise added by the forward process, which, as shown in Eq.\ref{eq:reverse_mean}, is equivalent to predicting imputation. Thus, our discussion focuses on imputation prediction, though it relies on noise prediction.

\subsection{Learning Feature Dependencies}
\label{subsec:fde}

In scenarios with partial blackouts, some features may be present at a given time step while others are missing. The FDE captures these relationships, enabling SADI to make accurate imputations. The FDE computes a new representation that captures joint time-series level relationships between features using self-attention, layer normalization, and a feed-forward network, focusing on the feature dimension to capture correlations.

Intutively, the FDE in our model uses a 1-D dilated convolution with a kernel size of $(1 \times 3)$ to extract locality information, iteratively updating the current time step. In each FDE layer, we increase the dilation by 1, expanding the receptive field to capture longer locality information, allowing both short and long-range patterns to be extracted. We then apply attention to capture joint time-series level feature correlations. By focusing attention on the feature dimension after the convolution, we effectively capture dependencies between different features based on time-series relationships. This combination of dilated convolutions and attention enables the FDE to extract local patterns while capturing global dependencies, allowing it to model complex interdependencies in time-series data more effectively.

The FDE takes $X$ as input, which we get from the noisy data, $X_t^{ta}$ and the observed data, $X_0^{co}$ as input after applying the missing mask, $M_0^{co}$ and a categorical positional encoding to it. The number of FDE layers is controlled by the hyperparameter $N_{FDE}$. 

\[FDE_n(X) = self\_attn(conv1\times 3(X^T, dilation=n)^T)\]
\begin{equation}
    \label{eq:fde}
    \hat{X} = 
    \begin{cases}
        FDE_n(X) & \text{if $n=1$} \\
        FDE_n(\hat{X}) & \text{if $1<n\le N_{FDE}$}
    \end{cases}
\end{equation}



\subsection{Learning Temporal Dependencies}
\label{subsec:gta}

To handle missing data in a partial blackout scenario, our model captures relationships between different time steps across all features using a gated temporal attention (GTA) block. This allows us to make accurate predictions even in scenarios with significant data gaps.

The gated temporal attention (GTA) block resembles the residual block architecture within DiffWave \cite{diffwave} and WaveNet \cite{oord2016wavenet} models. Both DiffWave and WaveNet are successful architectures for modeling sequential data, particularly for time-series and audio processing. Their residual block structures allow for efficient learning by ensuring that information flows through the network without vanishing gradients. While DiffWave and WaveNet use dilated convolutions to model long-range dependencies, we replace these with self-attention layers, introducing flexibility and adaptivity to capture non-local relationships, which is particularly useful for missing data scenarios. We also introduce positional encoding in the time dimension to signify it as a sequence for the self-attention mechanism. There is a Gated Linear Unit (GLU) activation applied to the outputs of the last self-attention layer, hence the name \emph{gated temporal attention}.

The GTA comprises multiple layers, controlled by the hyperparameter $N_{\text{GTA}}$, and takes $\hat{X}_{pos_1}$ and $X_{pos}^{co}$  as inputs, which we get after applying the missing mask $M_0^{co}$ and positional encoding to $\hat{X}$ (the output from FDE layers) and $X_0^{co}$ (observed data) respectively. GTA also takes the diffusion step embedding $t_{\text{emb}}$ as input. We use the positional encoding method from \cite{transformer-attn}. The GTA consists of two blocks, denoted as \( \text{GTA}^i() \) (\( i = \{1, 2\} \)), which output a hidden state (\( \Tilde{X} \)), attention weights (\( W_L \)), and a skip connection (\( \epsilon'_n \)) for the \( n \)-th layer. The skip connections from all \( N_{\text{GTA}} \) layers are aggregated to produce the imputation \( \epsilon_i \).

\begin{equation}
\label{eq:gta}
    \Tilde{X}, W_L, \epsilon'_n = 
    \begin{cases}
        GTA^i_n(\hat{X}_{pos_i}, X^{co}_{pos}, t_{emb}) & \text{if $n=1$} \\
        GTA^i_n(\Tilde{X}, X^{co}_{pos}, t_{emb}) & \text{otherwise}
    \end{cases}
\end{equation}

\begin{equation}
    \label{eq:eps1}
    \epsilon_1 = linear\left(\frac{\sum_{n=1}^{N_{GTA}}\epsilon'_n}{\sqrt{2}}\right)
\end{equation}

\subsection{Two-stage Imputation Process}
\label{subsec:two-stage}

After processing the data through multiple FDE and GTA layers to address feature and temporal dependencies, we introduce a second-stage GTA block to refine the predictions. This block specifically learns the temporal correlations revealed by the imputed values from the first stage. To preserve the original data's characteristics and enhance imputation quality, we reintroduce the original noisy data into the process. This reintroduction acts as a grounding mechanism, ensuring that the second-stage GTA block leverages both the transformed data and the raw input to produce more accurate and robust imputation results. Following the second GTA block, we obtain two interim imputation results: 
$\epsilon_1$ , derived from the first GTA block, and , $\epsilon_2$ refined by the second GTA block.

In practice, we observed that the second-stage imputation can sometimes degrade the model's accuracy. To address this, we introduce a weighted combination mechanism that learns the contribution of each interim imputation to the final result, thereby improving overall performance. Specifically, we combine the two interim outputs, $\epsilon_1$ and $\epsilon_2$, 
 to obtain the final imputation denoted as $\epsilon_{\theta}$ as depicted in Eq.~\ref{eq:wt}, where $\odot$ is the element-wise product.
The weighting coefficients, $\tilde{W_L}$, 
are acquired by applying missingness mask $M_0^{co}$ to $W_L (L,L)$ and a feed-forward network to project them to proper dimensions $(L, K)$ as shown in Eq.~\ref{eq:w_tilde}. 
\begin{equation}
\label{eq:w_tilde}
\Tilde{W}_L=sigmoid(linear(concat(W_L, M_0^{co})))
\end{equation}

\begin{equation}
    \label{eq:wt}
    \epsilon_{\theta} = (1-\tilde{W_L})\odot\epsilon_1 + \tilde{W_L}\odot\epsilon_2
\end{equation}

\subsection{Training and Sampling/Inference}
\label{subsec:train_samp}

We adopt the training methodology outlined in \cite{ho2020denoising}. This approach, illustrated in Algorithm~\ref{alg:training}, involves the uniform sampling of a diffusion step $t \in \{1,2,\ldots, T\}$ during each training iteration in Step~2. Next, in Step~3, we generate artificially missing data as the target for imputation using two different methods.
\begin{enumerate}
    \item \textbf{Randomly missing (RM):} In this approach, a percentage of data is randomly 
    missed during each epoch. In particular, we uniformly sample a time step $t$, and some feature $f$ where data is present and remove it. We repeat this process until a desired percentage of the data is masked, creating a controlled missing scenario for training the imputation model.
    
    \item \textbf{Mixed partial blackout (MPB):} This strategy begins by training the model using the random missing (RM) approach, allowing it to converge to a favorable local minimum. Once the model has achieved a good initial state, it is then fine-tuned by randomly introducing either partial blackout scenarios or random missing scenarios in each training iteration. We introduce blocks of missing data by following these steps:
    \begin{itemize}
        \item First, we randomly choose the number of features to remove, sampling from a uniform distribution $Uniform(1,\frac{n_{features}}{2})$.
        \item Next, we determine the number of consecutive time steps (or days) to remove, sampling from $Uniform(1,\frac{n_{time}}{2})$.
        \item We then uniformly sample which features to remove and randomly choose the starting time step for each missing block.
    \end{itemize}          
    
    By alternating between random missing and partial blackout scenarios in each epoch, the MPB method ensures that the model encounters a diverse range of missing patterns during training, enhancing its robustness across varied real-world missingness scenarios.
\end{enumerate}

\begin{algorithm}[t!]
    \caption{Training of our diffusion model}\label{alg:training}
    \textbf{Input:} Distribution of training data $X_0 \sim q(X_0)$, the number of iteration/epochs $E$, the list of noise levels $(\bar{\alpha}_1,\ldots,\bar{\alpha}_T)$\\
    \textbf{Output:}  Denoising function $\bm{\epsilon_{\theta}}$ 
    \begin{algorithmic}[1]
    
    \FOR {$i = 0$ to $E$}
        \STATE $t \sim Uniform(\{1,\ldots,T\})$
        \STATE Separate $X_0$ into conditional observations $X_0^{co}$ and imputation targets $X^{ta}_0$
        \STATE Noise $\epsilon=\mathcal{N}(0,\mathbf{I})$ with the same dimension as $X_0^{ta}$
        \STATE One step calculation to noisy targets at step $t$, $X_t^{ta}=\sqrt{\bar{\alpha}_t}X_0^{ta}+\sqrt{(1-\bar{\alpha}_t)}\epsilon $
        \STATE denoising prediction, $\epsilon_1, \epsilon_2, \epsilon_{\theta} =\bm{\epsilon_{\theta}} $ $(X^{ta}_t, X_0^{co},t)$
        \STATE Optimize the loss function for $\epsilon_{\theta}$, $\epsilon_1$, and $\epsilon_2$ according to Eq. (\ref{eq:loss}) and (\ref{eq:diff-loss}).
    \ENDFOR
    \end{algorithmic}
\end{algorithm}

In Step~5 of   Algorithm~\ref{alg:training}
we employ the closed form of the forward process to progress to any time step $t$ of diffusion. 
Our denoising function, $\bm{\epsilon_{\theta}}(X^{ta}_t, X_0^{co},t)$, is designed to predict the noise component $\epsilon$ that must be removed from $X_t^{ta}$ during the reverse process in Step~6. To optimize our model, we minimize the denoising loss function,  Eq.~\ref{eq:diff-loss}, which combines 
 $\epsilon_1, \epsilon_2$, and $\epsilon_{\theta}$.

\begin{equation}
\label{eq:diff-loss}
    loss = \frac{M^{ta}_0}{2N} \left(||\epsilon - \epsilon_{\theta}||_2^2 + \frac{(||\epsilon - \epsilon_1||_2^2 + ||\epsilon - \epsilon_2||_2^2)}{2} \right)
\end{equation}

In Eq.~\ref{eq:diff-loss}, $N$ is the number of imputation targets and $M^{ta}_0$ is the target mask where $1$ indicates the targets for imputation task and $0$ represents observed values and original missing data (without ground truth).

\begin{algorithm}[t!]
    \caption{Sampling process}\label{alg:sampling}
    \textbf{Input:} Data sample $X_0$, missingness mask $M_0^{co}$, total number of diffusion steps $T$, trained denoising function $\bm{\epsilon_{\theta}}$\\
    \textbf{Output} Imputed missing values $X_0^{ta}$ \ \\
    \vspace*{-0.15in}
    \begin{algorithmic}[1]   
    \STATE $X_0^{co} = \mbox{ observed values of } X_0$
    \STATE $X_{curr} = X_T^{ta}\sim \mathcal{N}(0,\mathbf{I})$ (same dimensions as $X_0$)
    \FOR {$t = T$ to $1$}
        \STATE $\epsilon_{\theta} = \bm{\epsilon_{\theta}}(X_{curr}, X_0^{co}, M_0^{co})$
        \STATE $\mu_{\theta} = \frac{1}{\sqrt{\alpha_t}}(X_{curr} - \frac{\beta_t}{\sqrt{1-\bar{\alpha}_t}}\epsilon_{\theta})$
        \STATE $\sigma_{\theta} = \frac{1-\bar{\alpha}_{t-1}}{1-\bar{\alpha}_t}\beta_t$; [Taken from \cite{ho2020denoising}]
        \IF{$t=0$}
            \STATE $X_{curr}=\mathcal{N}(\mu_{\theta}, \mathbf{I})$
        \ELSE
            \STATE $X_{curr}=\mathcal{N}(\mu_{\theta}, \sigma_{\theta}\mathbf{I})$
        \ENDIF
    \ENDFOR
    \STATE $X^{ta}_0 = X_{curr}$
    \STATE $X_0 = X_0^{co} \times M_0^{co} + X_0^{ta} \times (1-M_0^{co})$
    \end{algorithmic}
\end{algorithm}

During the inference phase shown in Algorithm~\ref{alg:sampling}, we generate imputed data for locations with missing values using a reverse diffusion process. This procedure is iterative, starting with pure Gaussian noise $X_T^{ta} \sim \mathcal{N}(0,\mathbf{I})$ at locations containing missing values. At each diffusion step $t$, we progressively remove some noise to produce the sample $X_{t-1}^{ta}$. To determine the noise to eliminate at diffusion step $t$, we utilize our proposed denoising function in Step~4.  
We first calculate the mean $\mu_{\theta}$ of $X_{t-1}^{ta}$ by removing the predicted noise $\epsilon_{\theta}$ from $X_t^{ta}$ in Step~5. The variance $\sigma_{\theta}$ is calculated in Step~6 following the formulation from \cite{ho2020denoising}. Step~7 to 10 show the formulation of $X_{t-1}^{ta}$ from the mean $\mu_{\theta}$ and the variance $\sigma_{\theta}$. Finally, we generate the output sample with predicted imputations in Step~14. We generate $50$ such samples and use their mean as the final imputation.

\section{Experiments}
\label{sec:experiments}

\begin{table*}[ht]
    \centering
    \resizebox{\textwidth}{!}{%
    \begin{tabular}{|c|c|c|c|c|c|c|c|}\hline
        \multirow{3}{1.4cm}{Datasets} &  \multirow{3}{1.2cm}{\# of Missing Features} & \multirow{3}{*}{MICE} & \multirow{3}{*}{BRITS} & \multirow{3}{*}{SAITS} & \multirow{3}{*}{CSDI} & \multirow{3}{*}{SADI-RM} & \multirow{3}{*}{SADI-MPB}\\
        & & & & & & &\\
        & & & & & & &\\\hline

        \multirow{6}{1.4cm}{AgAID} &  1 & 5.00e-03 $\pm$ 2.80e-03 & 8.71e-03 $\pm$ 4.80e-03 & 1.60e-03 $\pm$ 8.00e-03 &  3.63e-03 $\pm$ 2.13e-03 &  1.69e-04 $\pm$ 1.11e-04 & \textbf{1.09e-04 $\pm$ 6.49e-05} \\\cline{2-8}
         &  3 & 1.56e-02 $\pm$ 7.20e-03 & 1.35e-02 $\pm$ 6.81e-03 & 2.20e-03 $\pm$ 1.02e-03 & 4.33e-03 $\pm$ 1.30e-03 & 5.98e-04 $\pm$ 2.29e-04  & \textbf{2.93e-04 $\pm$ 1.04e-04} \\\cline{2-8}
         &  5 & 2.05e-02 $\pm$ 6.53e-03 & 1.37e-02 $\pm$ 6.42e-03 & 2.70e-03 $\pm$ 1.41e-03 & 5.12e-03 $\pm$ 1.86e-03 & 7.17e-04 $\pm$ 4.29e-04 & \textbf{3.27e-04  $\pm$ 1.46e-04} \\\cline{2-8}
         &  7 & 2.52e-02 $\pm$ 5.12e-03 & 1.50e-02 $\pm$ 3.13e-03 & 3.69e-03 $\pm$ 1.22e-03 & 6.15e-03 $\pm$ 2.33e-03 &  1.16e-03 $\pm$ 7.98e-04  &\textbf{5.36e-04 $\pm$ 3.29e-04} \\\cline{2-8}
         &  9 & 4.59e-02 $\pm$ 1.26e-02 & 2.56e-02 $\pm$ 1.00e-02 & 4.50e-03 $\pm$ 1.32e-03 & 5.74e-03 $\pm$ 1.66e-03 & 1.22e-03 $\pm$ 7.67e-04 &\textbf{5.96e-04 $\pm$ 3.50e-04} \\\cline{2-8}
         & 11 & 5.82e-02 $\pm$ 1.59e-02 & 3.35e-02 $\pm$ 1.13e-02 & 5.51e-03 $\pm$ 1.50e-03 & 8.51e-03 $\pm$ 2.11e-03 & 2.65e-03 $\pm$ 1.26e-03 & \textbf{1.25e-03 $\pm$ 6.06e-04} \\\hline\hline

         \multirow{6}{1.4cm}{Air Quality} &  1 & 2.80e-02 $\pm$ 1.00e-02 & 2.81e-02 $\pm$ 1.01e-02 & 2.80e-02 $\pm$ 1.01e-02 &  2.23e-03 $\pm$ 2.44e-03 & 1.74e-03 $\pm$ 1.91e-03  & \textbf{1.67e-03 $\pm$ 1.95e-03} \\\cline{2-8}
         & 3 & 2.07e-02 $\pm$ 1.83e-02 & 2.00e-02 $\pm$ 1.88e-02 & 2.06e-02 $\pm$ 1.83e-02 & 1.85e-03 $\pm$ 1.29e-03 & 1.04e-03 $\pm$ 4.01e-04  & \textbf{1.01e-03 $\pm$ 1.23e-03} \\\cline{2-8}
         &  5 & 1.66e-02 $\pm$ 7.22e-03 & 1.74e-02 $\pm$ 7.30e-03 & 1.66e-02 $\pm$ 7.11e-03 & 1.53e-03 $\pm$ 6.30e-04 & 1.09e-03 $\pm$ 5.90e-04  &  \textbf{1.07e-03 $\pm$ 4.06e-04} \\\cline{2-8}
         &  7 & 2.23e-02 $\pm$ 1.27e-02 & 2.20e-02 $\pm$ 1.27e-02 & 2.25e-02 $\pm$ 1.28e-02 & 1.54e-03 $\pm$ 4.00e-04 & 1.30e-03 $\pm$ 6.50e-04  & \textbf{1.09e-03 $\pm$ 9.00e-04} \\\cline{2-8}
         &  9 & 2.14e-02 $\pm$ 1.60e-02 & 2.03e-02 $\pm$ 1.51e-02 & 2.03e-02 $\pm$ 1.53e-02 & 1.47e-03 $\pm$ 5.92e-04 & 1.06e-03 $\pm$ 3.22e-04  & \textbf{1.04e-03 $\pm$ 2.15e-04} \\\cline{2-8}
         &  11 & 1.97e-02 $\pm$ 6.61e-03 &  1.92e-02 $\pm$ 6.20e-03 & 1.92e-02 $\pm$ 6.55e-03 & 1.24e-03 $\pm$ 2.11e-04 & 1.07e-03 $\pm$ 2.01e-04  & \textbf{9.80e-04 $\pm$ 5.07e-04} \\\hline\hline

         \multirow{6}{1.4cm}{Electricity} &  1 & 5.46e-01 $\pm$ 1.23e-01 & 8.74e-01 $\pm$ 4.09e-01 & 8.31e-01 $\pm$ 2.56e-01 & 3.56e-01 $\pm$ 4.98e-02 & 1.72e-01 $\pm$ 8.92e-02  & \textbf{1.70e-01 $\pm$ 8.90e-02} \\\cline{2-8}
         &  10 & 5.99e-01 $\pm$ 3.03e-02 & 1.03e-00 $\pm$ 4.74e-02 & 9.91e-01 $\pm$ 6.21e-02 & 6.31e-01 $\pm$ 5.90e-02 & 1.37e-01 $\pm$ 1.25e-02  & \textbf{1.07e-01 $\pm$ 1.18e-02} \\\cline{2-8}
         &  15 & 5.79e-01 $\pm$ 1.99e-02 &  9.39e-01 $\pm$ 7.70e-02 &  9.00e-01 $\pm$ 3.94e-02 & 5.70e-01 $\pm$ 8.32e-02 & 1.27e-01 $\pm$ 1.91e-02  & \textbf{1.17e-01 $\pm$ 1.90e-02} \\\cline{2-8}
         & 20 & 5.75e-01 $\pm$ 1.99e-02 & 9.23e-01 $\pm$ 4.29e-02 &  8.87e-01 $\pm$ 2.52e-02 & 5.86e-01 $\pm$ 4.77e-02 & 1.34e-01 $\pm$ 4.78e-02  & \textbf{1.25e-01 $\pm$ 3.47e-02} \\\cline{2-8}
         & 30 & 6.02e-01 $\pm$ 1.00e-02 & 9.97e-01 $\pm$ 1.34e-02 & 9.38e-01 $\pm$ 1.69e-02 & 5.61e-01 $\pm$ 3.53e-02 & 1.24e-01 $\pm$ 6.60e-03  & \textbf{1.14e-01 $\pm$ 5.54e-03} \\\cline{2-8}
         &  100 & 6.91e-01 $\pm$ 2.89e-02 & 1.00e-00 $\pm$ 2.02e-02 & 9.36e-01 $\pm$ 2.39e-02 & 4.39e-01 $\pm$ 2.09e-02 &  1.34e-01 $\pm$ 7.20e-03   & \textbf{1.02e-01 $\pm$ 7.10e-03} \\\hline\hline

         \multirow{5}{1.4cm}{NACSE} &  2 & 4.41e-03 $\pm$ 1.10e-03 & 7.12e-03 $\pm$ 3.02e-03 & 9.11e-03 $\pm$ 5.60e-03 & 1.31e-02 $\pm$ 1.87e-03 & 3.29e-03 $\pm$ 1.31e-03  &  \textbf{2.81e-03 $\pm$ 6.11e-04}\\\cline{2-8}
         &  10 & 5.13e-03 $\pm$ 8.00e-04 & 8.20e-03 $\pm$ 2.11e-03 & 1.09e-02 $\pm$ 2.77e-03 & 1.50e-02 $\pm$ 1.56e-03 & 5.93e-03 $\pm$ 2.53e-03  & \textbf{4.88e-03 $\pm$ 8.94e-04} \\\cline{2-8}
         &  50 & 7.01e-03 $\pm$ 9.00e-04 & 8.33e-03 $\pm$ 9.01e-04 & 1.00e-02 $\pm$ 1.32e-03 & 1.49e-02 $\pm$ 8.85e-04 & 5.35e-03 $\pm$ 1.04e-03  & \textbf{4.80e-03 $\pm$ 7.55e-04} \\\cline{2-8}
         &  90 & 1.16e-02 $\pm$ 8.23e-04 & 1.17e-02 $\pm$ 1.00e-03 & 1.09e-02 $\pm$ 1.01e-03 & 1.58e-02 $\pm$ 6.23e-04 & 6.33e-03 $\pm$ 1.07e-03  & \textbf{5.29e-03 $\pm$ 6.39e-04}  \\\cline{2-8}
         &  100 & 1.17e-02 $\pm$ 1.22e-03 & 1.17e-02 $\pm$ 1.30e-03 &  1.06e-02 $\pm$ 1.10e-03 & 1.43e-02 $\pm$ 6.72e-04 & 6.08e-03 $\pm$ 9.33e-04   & \textbf{5.08e-03 $\pm$ 4.88e-04}  \\\hline
    \end{tabular}}
    \caption{MSE (with 95\% confidence interval) is calculated by averaging 20 inference trials (length of missing time period = 10 (Air Quality) and 30 (for the rest)).}
    \label{tab:real_pbm}
\end{table*}

\begin{table}[h!]
    \centering
    \resizebox{0.48\textwidth}{!}{%
    \begin{tabular}{|c|c|c|c|c|c|}\hline
        \multirow{3}{*}{Data}  & \multirow{3}{*}{MF}  & \multirow{3}{*}{CSDI} & \multirow{3}{*}{SADI-RM} & \multirow{3}{*}{SADI-MPB}\\
        &    &  &  &\\
        &    &  &  &\\\hline
        \multirow{6}{*}{AG} &   1 & 7.26e-03 $\pm$ 2.08e-03 & 1.79e-03 $\pm$ 6.40e-04 & \textbf{1.09e-03 $\pm$ 5.23e-04} \\\cline{2-5}
         &   3 & 8.25e-03 $\pm$ 1.15e-03 & 2.86e-03 $\pm$ 4.56e-04 & \textbf{2.93e-04 $\pm$ 3.36e-04} \\\cline{2-5}
         &   5 & 8.47e-03 $\pm$ 1.12e-03 & 2.91e-03 $\pm$ 5.46e-04 &  \textbf{3.27e-04 $\pm$ 3.22e-04} \\\cline{2-5}
         &   7 & 9.49e-04 $\pm$ 1.08e-04 & 3.46e-03 $\pm$ 6.82e-04 & \textbf{5.36e-04 $\pm$ 4.06e-04} \\\cline{2-5}
         &   9 & 9.74e-03 $\pm$ 1.04e-03 & 3.90e-03 $\pm$ 8,03e-04 & \textbf{5.96e-04 $\pm$ 5.71e-04} \\\cline{2-5}
         &   11 & 1.13e-02 $\pm$ 1.06e-03 & 5.15e-03 $\pm$ 1.06e-03 & \textbf{1.25e-03 $\pm$ 6.73e-04} \\\hline

         \multirow{6}{*}{AQ}  & 1 & 5.61e-03 $\pm$ 7.60e-04 & 3.70e-03 $\pm$ 1.19e-03 & \textbf{3.27e-03 $\pm$ 1.69e-03} \\\cline{2-5}
         &   3 & 8.47e-03 $\pm$ 4.50e-03 &  5.50e-03 $\pm$ 2.91e-03 & \textbf{4.20e-03 $\pm$ 3.48e-03} \\\cline{2-5}
         &   5 & 6.92e-03 $\pm$ 1.04e-03 & 3.58e-03 $\pm$ 4.56e-04 & \textbf{1.09e-03 $\pm$ 2.31e-04} \\\cline{2-5}
         &   7 & 8.30e-03 $\pm$ 9.18e-04 & 5.14e-03 $\pm$ 1.53e-03 & \textbf{4.74e-03 $\pm$ 7.77e-03} \\\cline{2-5}
         &   9 & 7.48e-03 $\pm$ 1.88e-03 & 4.57e-03 $\pm$ 1.12e-03 & \textbf{2.89e-03 $\pm$ 1.84e-03} \\\cline{2-5}
         &  11 & 8.31e-03 $\pm$ 1.35e-03 & 4.82e-03 $\pm$ 1.33e-03 & \textbf{3.04e-03 $\pm$ 1.17e-03} \\\hline

         \multirow{6}{*}{EL}  & 1 & 7.40e-01 $\pm$ 9.03e-03 & 1.55e-01 $\pm$ 1.29e-02 &  \textbf{5.31e-02 $\pm$ 1.22e-02}  \\\cline{2-5}
         &   10 & 7.26e-01 $\pm$ 2.26e-03 & 1.64e-01 $\pm$ 4.63e-03 & \textbf{5.32e-02 $\pm$ 1.38e-03} \\\cline{2-5}
         &   15 & 7.36e-01 $\pm$ 3.64e-03 & 1.62e-01 $\pm$ 5.14e-03 & \textbf{5.10e-02 $\pm$ 2.28e-03} \\\cline{2-5}
         &   20 & 7.46e-01 $\pm$ 3.14e-03 & 1.63e-01 $\pm$ 3.20e-03 & \textbf{5.30e-02 $\pm$ 2.51e-03} \\\cline{2-5}
         &   30 & 7.72e-01 $\pm$ 3.10e-03 & 1.63e-01 $\pm$ 1.46e-03 & \textbf{5.21e-02 $\pm$ 2.20e-03}  \\\cline{2-5}
         &   100 & 7.94e-01 $\pm$ 4.22e-03 & 1.70e-01 $\pm$ 2.17e-03 & \textbf{5.78e-02 $\pm$ 5.79e-03} \\\hline

         \multirow{6}{*}{NE}  & 2 & 2.11e-02 $\pm$ 2.71e-03 & 1.04e-02 $\pm$ 9.65e-04 & \textbf{3.29e-03 $\pm$ 1.34e-03} \\\cline{2-5}
         &  10 & 2.12e-02 $\pm$ 2.74e-03 & 1.21e-02 $\pm$ 1.38e-03 & \textbf{5.93e-03 $\pm$ 1.31e-03} \\\cline{2-5}
         &   50 & 2.20e-02 $\pm$ 1.10e-03 & 1.27e-02 $\pm$ 8.23e-04 & \textbf{5.35e-03 $\pm$ 7.15e-04} \\\cline{2-5}
         &   90 & 2.20e-02 $\pm$ 1.67e-03 & 1.28e-02 $\pm$ 8.93e-04 & \textbf{6.33e-03 $\pm$ 7.91e-04} \\\cline{2-5}
         &   100 & 1.98e-02 $\pm$ 1.29e-03 & 1.20e-02 $\pm$ 8.74e-04 & \textbf{6.08e-03 $\pm$ 9.01e-04} \\\hline
    \end{tabular}}
    \caption{Abbreviations: Data = Datasets, AG = AgAID, AQ = Air Quality, EL = Electricity, NE = NACSE, and MF = Number of missing features. CRPS (with 95\% confidence interval) is calculated by averaging 20 inference trials (length of missing time period = 10 (Air Quality) and 30 (for the rest)).}
    \label{tab:real_pbm_crps}
\end{table}

\begin{table*}[ht]
    \centering
    \resizebox{0.90\textwidth}{!}{%
    \begin{tabular}{|c|c|c|c|c|c|c|}\hline
        \multirow{2}{*}{Datasets} &  \multirow{2}{2.5cm}{\# of Missing Features} & \multicolumn{4}{|c|}{MSE $\pm$ 95\% CI} \\\cline{3-6}
        &  & SADI & No FDE & No 2nd block & No wt. comb.\\\hline
        
        \multirow{3}{*}{AgAID} &  3 & \textbf{2.93e-04 $\pm$ 1.04e-04} & 1.43e-03 $\pm$ 9.00e-04 & 1.12e-03 $\pm$ 3.10e-04 & 7.81e-04 $\pm$ 5.62e-04 \\\cline{2-6}
      &  7 & \textbf{5.36e-04 $\pm$ 3.29e-04} & 1.77e-03 $\pm$ 1.72e-03 & 1.53e-03 $\pm$ 3.70e-04 & 5.41e-03 $\pm$ 9.40e-03\\\cline{2-6}
      &  11 & \textbf{1.25e-03 $\pm$ 6.06e-04} & 1.64e-03 $\pm$ 8.00e-04 & 2.10e-03 $\pm$ 4.30e-04 & 6.75e-03 $\pm$ 3.87e-03\\\hline\hline

        \multirow{3}{*}{NACSE} &  10 & \textbf{4.88e-03 $\pm$ 6.11e-04} & 3.74e-02 $\pm$ 2.04e-02 & 4.10e-02 $\pm$ 2.58e-02 & 3.95e-02 $\pm$ 1.61e-02\\\cline{2-6}
      &  50 & \textbf{4.80e-03 $\pm$ 7.50e-04} & 7.81e-02 $\pm$ 1.49e-02 &  7.79e-02 $\pm$ 2.34e-02 & 6.48e-02 $\pm$ 1.81e-02\\\cline{2-6}
      &  100 & \textbf{5.08e-03 $\pm$ 4.80e-04} & 8.60e-02 $\pm$ 1.21e-02 &  7.49e-02 $\pm$ 1.78e-02 & 6.36e-02 $\pm$ 9.12e-03\\\hline\hline

      \multirow{3}{*}{Air Quality} &  5 & \textbf{1.07e-03 $\pm$ 4.10e-04} & 3.42e-03 $\pm$ 1.50e-03 & 4.18e-03 $\pm$ 2.78e-03 & 2.21e-03 $\pm$ 1.14e-03 \\\cline{2-6}
      &  9 & \textbf{1.04e-03 $\pm$ 2.20e-04} & 4.76e-03 $\pm$ 3.82e-03 & 2.48e-03 $\pm$ 8.91e-04 & 1.15e-03 $\pm$ 9.00e-04 \\\cline{2-6}
      &  11 & \textbf{9.80e-04 $\pm$ 5.11e-04} & 3.51e-03 $\pm$ 9.50e-04 & 2.35e-03 $\pm$ 2.10e-04 & 3.88e-03 $\pm$ 2.41e-03\\\hline\hline

      \multirow{3}{*}{Electricity} &  10 & \textbf{1.07e-01 $\pm$ 1.18e-02} & 9.58e+00 $\pm$ 3.88e-01 & 7.09e-01 $\pm$ 6.06e-02 & 2.10e+00 $\pm$ 1.49e-01 \\\cline{2-6}
      &  30 & \textbf{1.14e-01 $\pm$ 5.54e-03} & 9.15e+00 $\pm$ 2.25e-01 & 7.37e-01 $\pm$ 4.32e-02 & 2.15e+00 $\pm$ 9.94e-02\\\cline{2-6}
      &  100 & \textbf{1.02e-01 $\pm$ 7.10e-03} & 8.10e+00 $\pm$ 4.78e-02 &  7.79e-01 $\pm$ 4.54e-03  & 2.06e+00 $\pm$  5.85e-04\\\hline
    \end{tabular}}
    \caption{Ablation Study of the three core components of SADI: MSE (with 95\% confidence interval) is calculated by averaging across three training runs each running 20 inference trials.}
    \label{tab:ablation}
\end{table*}

\subsection{Time-series Datasets}
\label{subsec:real}

The first dataset is a grape cultivar cold hardiness dataset from \textbf{AgAID}, which measures grape plant characteristics such as 
its resistance to cold weather 
along with a number of environmental factors at regular intervals\footnote{https://shorturl.at/zk0Ia}. This dataset has 34.41\% missing data.
It covers dormant seasons from September 7 to May 15 with $\mathbf{21}$ features and $\mathbf{252}$ time steps. It spans 34 seasons (1988 to 2022), with the last 2 seasons set aside for testing.

\textbf{Air Quality} is a popular dataset considered in \citep{geo-sensory-miss} among others. In accordance with prior research  \citep{ode-score, brits, csdi}, we utilize hourly PM2.5 measurements from $\mathbf{36}$ stations (features) located in Beijing, covering a period of 12 months. We aggregate these measurements into time series, each consisting of $\mathbf{36}$ consecutive time steps. This dataset exhibits an approximate $13\%$ rate of missing values.

Another widely known dataset is the \textbf{Electricity} Load Diagram from the public UCI machine learning repository \cite{uci}. It comprises electricity consumption data, measured in kilowatt-hours (kWh), gathered from $\mathbf{370}$ clients, treated as $\mathbf{370}$ features, at $15$-minute intervals for $\mathbf{100}$ time steps.
This dataset has no original missing data. There are 48 months worth of data.  
We designate the first 10 months of data as the test set, the subsequent 10 months as the validation set, and the remaining data as the training set, the same as \cite{saits}.

The last dataset examined in this study consists of temperature data sourced from the Northwest Alliance for Computational Science \& Engineering (\textbf{NACSE}) PRISM climate data\footnote{https://shorturl.at/Aor04}. 
It has the maximum and minimum temperatures recorded daily in $176$ weather stations, 
which makes up a total of 
$\mathbf{352}$ features.  The time-series spans a year, specifically $\mathbf{366}$ days. The dataset contains data for a total of 11 years and has 23.72\% originally missing data. For our experimental setup, we reserve the last 2 years for testing.

\subsection{Experimental Setup}
We evaluate the performance of SADI against several benchmark models: CSDI \citep{csdi} (a conditional diffusion-based model), BRITS \citep{brits} (a bidirectional RNN-based autoregressive model), SAITS \citep{saits} (a self-attention-based model), and MICE \citep{mice} (an iterative linear regression-based model). The comparison is conducted on datasets under a partial blackout scenario. Additionally, we compare the two training strategies of our model, denoted as SADI-RM and SADI-MPB, as shown in Table~\ref{tab:real_pbm}.

We train the models once and test them $20$ times on the test set in different missingness settings (different ground truths) under partial blackout. For each test trial, we uniformly select which features are missing and select two blocks to be missing for $10$ (in the case of the Air Quality dataset) or $30$ (in the case of other datasets) consecutive time steps. For CSDI and SADI, we generated $50$ predicted samples to approximate the probability distribution of the missing data. We used the mean for SADI and the median for CSDI (as proposed in their paper) for the final prediction. The other three models make point predictions for imputation.



To assess the performance of SADI, we rely on two key metrics - the mean squared error (MSE) and the Continuous Ranked Probability Score (CRPS) \cite{CRPS} - with $95\%$ confidence intervals. CRPS is a statistical metric used to measure the accuracy of probabilistic forecasts. This metric measures the difference between the cumulative distribution function (CDF) of the predicted values and the pointwise ground truth values. It is calculated as the integral of the squared differences between the predicted CDF and the actual values. If, say, $F$ is a function that predicts a distribution and the ground truth is $y$, then the CRPS formulation is given in Eq.~\ref{eq:crps}. A lower CRPS value indicates a more accurate prediction. Since CRPS is a measure of performance for generative models, we use this metric only for CSDI and SADI.

\begin{equation}
\label{eq:crps}
    CRPS(F,y) = \int{(F(x)-\mathbf{1}_{x\ge y})^2}dx
\end{equation}

Tables~\ref{tab:real_pbm} and \ref{tab:real_pbm_crps} show the MSE and CRPS results with $95\%$ confidence intervals for the four real-world datasets. Here, we can observe that SADI outperforms the other models in MSE and CRPS in partial blackout scenarios. 
Among the two variants of training methods, the mixed partial blackouts  approach (SADI-MPB) shows superior performance compared to the model trained solely with random missing scenarios (SADI-RM). This highlights the effectiveness of the mixed training strategy in enhancing the model's robustness to more challenging data gaps.
In these experiments, we have observed that CSDI (code taken from one of the author's GitHub repository\footnote{https://github.com/ermongroup/CSDI}) requires a huge amount of GPU memory when dealing with high dimensional data such as - Electricity and NACSE datasets. For these two datasets, we had to reduce the number of channels to $8$ because of our GPU constraints, which may have had some negative effect on its performance shown in 
Tables~\ref{tab:real_pbm} and \ref{tab:real_pbm_crps}. 
Some synthetic data experiments have been reported in the supplementary materials. These experiments revealed that our model outperformed other models across all datasets except one, which lacked interdependency among the features. This observation suggests that our model excels in scenarios where there is significant interdependency among the features.

\subsection{Ablation Study}
\label{subsec:ablation}
Our model, SADI, has three core features: (1) the FDE (feature dependency encoder) block that models feature inter-correlations, (2) the two-stage imputation process, and (3) the weighted combination of the two intermediate imputations. Now, we will do an ablation study to show the impact of these three design decisions.
\begin{itemize}
    \item \textbf{SADI}: The SADI model with all of its components.
    \item \textbf{No FDE}: SADI model without the FDE component.
    \item \textbf{No 2nd block}: SADI model after removing the second stage of imputation. Instead of having two separate $N_{GTA}$ layers for each block, we now have a single block with $2 \times N_{GTA}$ layers. It takes the first stage's output as the final imputation.
    \item \textbf{No wt. comb.}: SADI model without the weighted combination of two blocks. It takes the prediction of the second stage as the final imputation.
\end{itemize}

Table~\ref{tab:ablation} shows the ablation results for all three design choices on the four datasets. For every model, the training is done $3$ times, and each time, there were $20$ inference runs with different missing scenarios. Table~\ref{tab:ablation} reports the MSE with $95\%$ confidence interval averaging across all training and inference runs.
We observe that with all of its components, SADI outperforms the versions that lack at least one of the core components in all datasets. We can see that the FDE component is particularly important for the NACSE and Electricity datasets because the {\em no FDE} model performs a lot worse than SADI compared to the other
two ablation models. It means that these two datasets have high correlation among the features.



\section{Discussion and Conclusion}
\label{sec:future_works}
In this paper, we addressed the multivariate time-series imputation, which is a critical problem in many domains. While previous research has made significant strides in this area, much of it has focused on limited missing data scenarios, leaving a gap in addressing more realistic and complex patterns of missingness. To close this gap, we introduced a more general pattern of missingness called \textbf{partial blackouts} that includes a wide variety of missing data patterns found in the real world. 

A common drawback in many existing methods is the lack of explicit modeling of feature dependencies at the time-series level, leading to suboptimal imputation quality. Among state-of-the-art models, only CSDI addresses this issue. Our proposed SADI explicitly captures both feature and temporal correlations, offering robust imputation across diverse features. Its two-step process improves quality by considering the influence of initial imputations, though the second stage may not always outperform the first. To address this, we introduced a learnable dynamic weighting mechanism, which ablation studies show is critical to our method's success.


In Section~\ref{sec:experiments}, we demonstrated that SADI outperforms state-of-the-art models in both MSE and CRPS metrics. Among its variants, the mixed partial blackout training version, SADI-MPB, proved most effective. This approach initializes SADI to a strong baseline and fine-tunes it for partial blackout scenarios, enhancing robustness and performance in handling significant data gaps, solidifying SADI as a leading time-series imputation method.

Moreover, SADI requires less GPU memory than CSDI for training and inference, making it suitable for diverse applications. While running CSDI on large datasets like Electricity and NACSE required reducing the number of channels due to memory constraints, SADI handled the same capacity without such adjustments.

In summary, this paper introduces the ``partial blackout" framework for multivariate time-series imputation, offering a realistic evaluation of models. SADI provides a robust, scalable solution, advancing the state of the art with high-quality imputations that capture complex feature and temporal dependencies.

\section*{Acknowledgements}
This research was supported by \textbf{NSF and USDA-NIFA under the AI Institute: Agricultural AI for Transforming Workforce and Decision Support (AgAID\footnote{https://agaid.org/}) award No. 2021-67021-35344}. We thank Lynn Mills, Alan Kawakami, Marcus Keller, and his viticulture team at Washington State University for providing their Grape Coldhardiness data. We are grateful to Chris Daly and Dylan Keon for providing us with the NACSE temperature data.

\bibliography{references}

\begin{thebibliography}{33}
\providecommand{\natexlab}[1]{#1}

\bibitem[{Alcaraz and Strodthoff(2022)}]{sssd}
Alcaraz, J. M.~L.; and Strodthoff, N. 2022.
\newblock Diffusion-based time series imputation and forecasting with structured state space models.
\newblock \emph{arXiv preprint arXiv:2208.09399}.

\bibitem[{Cao et~al.(2018)Cao, Wang, Li, Zhou, Li, and Li}]{brits}
Cao, W.; Wang, D.; Li, J.; Zhou, H.; Li, L.; and Li, Y. 2018.
\newblock Brits: Bidirectional recurrent imputation for time series.
\newblock \emph{Advances in neural information processing systems}, 31.

\bibitem[{Che et~al.(2016)Che, Purushotham, Cho, Sontag, and Liu}]{gru-d}
Che, Z.; Purushotham, S.; Cho, K.; Sontag, D.~A.; and Liu, Y. 2016.
\newblock Recurrent Neural Networks for Multivariate Time Series with Missing Values.
\newblock \emph{CoRR}, abs/1606.01865.

\bibitem[{Chen et~al.(2021)Chen, Zhang, Zen, Weiss, Norouzi, and Chan}]{wavegrad}
Chen, N.; Zhang, Y.; Zen, H.; Weiss, R.~J.; Norouzi, M.; and Chan, W. 2021.
\newblock WaveGrad: Estimating Gradients for Waveform Generation.
\newblock In \emph{International Conference on Learning Representations}.

\bibitem[{Cini, Marisca, and Alippi(2021)}]{grin}
Cini, A.; Marisca, I.; and Alippi, C. 2021.
\newblock Multivariate Time Series Imputation by Graph Neural Networks.
\newblock \emph{CoRR}, abs/2108.00298.

\bibitem[{Cui et~al.(2019)Cui, Bucur, Groot, and Heskes}]{Cui2019}
Cui, R.; Bucur, I.~G.; Groot, P.; and Heskes, T. 2019.
\newblock A novel Bayesian approach for latent variable modeling from mixed data with missing values.
\newblock \emph{Statistics and Computing}, 29(5): 977–993.

\bibitem[{Du, C{\^o}t{\'e}, and Liu(2023)}]{saits}
Du, W.; C{\^o}t{\'e}, D.; and Liu, Y. 2023.
\newblock Saits: Self-attention-based imputation for time series.
\newblock \emph{Expert Systems with Applications}, 219: 119619.

\bibitem[{Dua and Graff(2017)}]{uci}
Dua, D.; and Graff, C. 2017.
\newblock UCI Machine Learning Repository.

\bibitem[{Fortuin et~al.(2020)Fortuin, Baranchuk, R{\"a}tsch, and Mandt}]{gp-vae}
Fortuin, V.; Baranchuk, D.; R{\"a}tsch, G.; and Mandt, S. 2020.
\newblock Gp-vae: Deep probabilistic time series imputation.
\newblock In \emph{International conference on artificial intelligence and statistics}, 1651--1661. PMLR.

\bibitem[{Ho, Jain, and Abbeel(2020)}]{ho2020denoising}
Ho, J.; Jain, A.; and Abbeel, P. 2020.
\newblock Denoising diffusion probabilistic models.
\newblock \emph{Advances in Neural Information Processing Systems}, 33: 6840--6851.

\bibitem[{Kong et~al.(2020)Kong, Ping, Huang, Zhao, and Catanzaro}]{diffwave}
Kong, Z.; Ping, W.; Huang, J.; Zhao, K.; and Catanzaro, B. 2020.
\newblock Diffwave: A versatile diffusion model for audio synthesis.
\newblock \emph{arXiv preprint arXiv:2009.09761}.

\bibitem[{Liu et~al.(2019)Liu, Yu, Zheng, Zhan, and Yue}]{naomi}
Liu, Y.; Yu, R.; Zheng, S.; Zhan, E.; and Yue, Y. 2019.
\newblock {NAOMI:} Non-Autoregressive Multiresolution Sequence Imputation.
\newblock \emph{CoRR}, abs/1901.10946.

\bibitem[{Luo et~al.(2018)Luo, Cai, ZHANG, Xu, and xiaojie}]{grui-gan}
Luo, Y.; Cai, X.; ZHANG, Y.; Xu, J.; and xiaojie, Y. 2018.
\newblock Multivariate Time Series Imputation with Generative Adversarial Networks.
\newblock In Bengio, S.; Wallach, H.; Larochelle, H.; Grauman, K.; Cesa-Bianchi, N.; and Garnett, R., eds., \emph{Advances in Neural Information Processing Systems}, volume~31. Curran Associates, Inc.

\bibitem[{Luo et~al.(2019)Luo, Zhang, Cai, and Yuan}]{e2gan}
Luo, Y.; Zhang, Y.; Cai, X.; and Yuan, X. 2019.
\newblock E²GAN: End-to-End Generative Adversarial Network for Multivariate Time Series Imputation.
\newblock In \emph{Proceedings of the Twenty-Eighth International Joint Conference on Artificial Intelligence, {IJCAI-19}}, 3094--3100. International Joint Conferences on Artificial Intelligence Organization.

\bibitem[{Matheson and Winkler(1976)}]{CRPS}
Matheson, J.~E.; and Winkler, R.~L. 1976.
\newblock Scoring Rules for Continuous Probability Distributions.
\newblock \emph{Management Science}, 22(10): 1087--1096.

\bibitem[{Miao et~al.(2021)Miao, Wu, Wang, Gao, Mao, and Yin}]{miao2021generative}
Miao, X.; Wu, Y.; Wang, J.; Gao, Y.; Mao, X.; and Yin, J. 2021.
\newblock Generative Semi-supervised Learning for Multivariate Time Series Imputation.
\newblock In \emph{AAAI}.

\bibitem[{Nichol and Dhariwal(2021)}]{nichol2021improved}
Nichol, A.~Q.; and Dhariwal, P. 2021.
\newblock Improved denoising diffusion probabilistic models.
\newblock In \emph{International Conference on Machine Learning}, 8162--8171. PMLR.

\bibitem[{Oord et~al.(2016)Oord, Dieleman, Zen, Simonyan, Vinyals, Graves, Kalchbrenner, Senior, and Kavukcuoglu}]{oord2016wavenet}
Oord, A. v.~d.; Dieleman, S.; Zen, H.; Simonyan, K.; Vinyals, O.; Graves, A.; Kalchbrenner, N.; Senior, A.; and Kavukcuoglu, K. 2016.
\newblock Wavenet: A generative model for raw audio.
\newblock \emph{arXiv preprint arXiv:1609.03499}.

\bibitem[{Park et~al.(2022)Park, M\"{u}ller, Arora, Faybishenko, Pastorello, Varadharajan, Sahu, and Agarwal}]{long-term-Park2022}
Park, J.; M\"{u}ller, J.; Arora, B.; Faybishenko, B.; Pastorello, G.; Varadharajan, C.; Sahu, R.; and Agarwal, D. 2022.
\newblock Long-term missing value imputation for time series data using deep neural networks.
\newblock \emph{Neural Computing and Applications}.

\bibitem[{Rasul et~al.(2021)Rasul, Seward, Schuster, and Vollgraf}]{time-grad-25}
Rasul, K.; Seward, C.; Schuster, I.; and Vollgraf, R. 2021.
\newblock Autoregressive Denoising Diffusion Models for Multivariate Probabilistic Time Series Forecasting.
\newblock \emph{CoRR}, abs/2101.12072.

\bibitem[{Rubanova, Chen, and Duvenaud(2019)}]{latent-ode}
Rubanova, Y.; Chen, R. T.~Q.; and Duvenaud, D.~K. 2019.
\newblock Latent Ordinary Differential Equations for Irregularly-Sampled Time Series.
\newblock In Wallach, H.; Larochelle, H.; Beygelzimer, A.; d\textquotesingle Alch\'{e}-Buc, F.; Fox, E.; and Garnett, R., eds., \emph{Advances in Neural Information Processing Systems}, volume~32. Curran Associates, Inc.

\bibitem[{Shadbahr et~al.(2022)Shadbahr, Roberts, Stanczuk, Gilbey, Teare, Dittmer, Thorpe, Torne, Sala, Lio et~al.}]{shadbahr2022classification}
Shadbahr, T.; Roberts, M.; Stanczuk, J.; Gilbey, J.; Teare, P.; Dittmer, S.; Thorpe, M.; Torne, R.~V.; Sala, E.; Lio, P.; et~al. 2022.
\newblock Classification of datasets with imputed missing values: Does imputation quality matter?
\newblock \emph{arXiv preprint arXiv:2206.08478}.

\bibitem[{Silva et~al.(2012)Silva, Moody, Scott, Celi, and Mark}]{Silva2012-cb}
Silva, I.; Moody, G.; Scott, D.~J.; Celi, L.~A.; and Mark, R.~G. 2012.
\newblock Predicting in-hospital mortality of {ICU} patients: The {PhysioNet/computing} in cardiology challenge 2012.
\newblock \emph{Comput. Cardiol. (2010)}, 39: 245--248.

\bibitem[{Sohl-Dickstein et~al.(2015)Sohl-Dickstein, Weiss, Maheswaranathan, and Ganguli}]{sohl2015deep}
Sohl-Dickstein, J.; Weiss, E.; Maheswaranathan, N.; and Ganguli, S. 2015.
\newblock Deep unsupervised learning using nonequilibrium thermodynamics.
\newblock In \emph{International Conference on Machine Learning}, 2256--2265. PMLR.

\bibitem[{Song et~al.(2020)Song, Sohl-Dickstein, Kingma, Kumar, Ermon, and Poole}]{ode-score}
Song, Y.; Sohl-Dickstein, J.; Kingma, D.~P.; Kumar, A.; Ermon, S.; and Poole, B. 2020.
\newblock Score-based generative modeling through stochastic differential equations.
\newblock \emph{arXiv preprint arXiv:2011.13456}.

\bibitem[{Tashiro et~al.(2021)Tashiro, Song, Song, and Ermon}]{csdi}
Tashiro, Y.; Song, J.; Song, Y.; and Ermon, S. 2021.
\newblock {CSDI:} Conditional Score-based Diffusion Models for Probabilistic Time Series Imputation.
\newblock \emph{CoRR}, abs/2107.03502.

\bibitem[{van Buuren and Groothuis-Oudshoorn(2011)}]{mice}
van Buuren, S.; and Groothuis-Oudshoorn, K. 2011.
\newblock mice: Multivariate Imputation by Chained Equations in R.
\newblock \emph{Journal of Statistical Software}, 45(3): 1–67.

\bibitem[{Vaswani et~al.(2017)Vaswani, Shazeer, Parmar, Uszkoreit, Jones, Gomez, Kaiser, and Polosukhin}]{transformer-attn}
Vaswani, A.; Shazeer, N.; Parmar, N.; Uszkoreit, J.; Jones, L.; Gomez, A.~N.; Kaiser, {\L}.; and Polosukhin, I. 2017.
\newblock Attention is all you need.
\newblock \emph{Advances in neural information processing systems}, 30.

\bibitem[{Vidotto, Vermunt, and Van~Deun(2018)}]{Vidotto2018}
Vidotto, D.; Vermunt, J.~K.; and Van~Deun, K. 2018.
\newblock Bayesian Latent Class Models for the Multiple Imputation of Categorical Data.
\newblock \emph{Methodology}, 14(2): 56–68.

\bibitem[{Vidotto, Vermunt, and Van~Deun(2019)}]{Vidotto2019}
Vidotto, D.; Vermunt, J.~K.; and Van~Deun, K. 2019.
\newblock Multiple imputation of longitudinal categorical data through bayesian mixture latent Markov models.
\newblock \emph{Journal of Applied Statistics}, 47(10): 1720–1738.

\bibitem[{Xiang et~al.(2023)Xiang, Yang, Huang, and Tong}]{3daware}
Xiang, J.; Yang, J.; Huang, B.; and Tong, X. 2023.
\newblock 3D-aware Image Generation using 2D Diffusion Models.
\newblock \emph{arXiv preprint arXiv:2303.17905}.

\bibitem[{Yi et~al.(2016)Yi, Zheng, Zhang, and Li}]{geo-sensory-miss}
Yi, X.; Zheng, Y.; Zhang, J.; and Li, T. 2016.
\newblock ST-MVL: Filling Missing Values in Geo-Sensory Time Series Data.
\newblock In \emph{Proceedings of the Twenty-Fifth International Joint Conference on Artificial Intelligence}, IJCAI'16, 2704–2710. AAAI Press.
\newblock ISBN 9781577357704.

\bibitem[{Zhang et~al.(2021)Zhang, Xiao, Zhou, Zhu, and Amos}]{Zhang2021-pw}
Zhang, Z.; Xiao, X.; Zhou, W.; Zhu, D.; and Amos, C.~I. 2021.
\newblock False positive findings during genome-wide association studies with imputation: influence of allele frequency and imputation accuracy.
\newblock \emph{Hum. Mol. Genet.}, 31(1): 146--155.

\end{thebibliography}



\end{document}